\documentclass{article}

\usepackage{CJKutf8}
\usepackage{longcat_style}
\usepackage{adjustbox}
\usepackage[utf8]{inputenc} %
\usepackage[T1]{fontenc}    %
\usepackage{newunicodechar}
\usepackage{hyperref}       %
\usepackage{xcolor}
\usepackage[normalem]{ulem} %
\hypersetup{
    colorlinks=true,      %
    linkcolor=blue,      %
    urlcolor=blue,       %
    citecolor=blue,      %
    linkbordercolor=blue, %
    urlbordercolor=blue,
    citebordercolor=blue,
    pdfborderstyle={/S/U/W 1}, %
}
\usepackage{float}
\usepackage{url}            %
\usepackage{booktabs}       %
\usepackage{amsfonts}       %
\usepackage{nicefrac}       %
\usepackage{microtype}      %
\usepackage{lipsum}		%
\usepackage{graphicx}
\usepackage{natbib}
\usepackage{doi}
\usepackage{amsmath}
\usepackage{algorithm}
\usepackage{algpseudocode}
\usepackage{amssymb} %
\usepackage{xspace}
\usepackage{enumitem}
\usepackage{multirow}
\usepackage{subcaption} 
\usepackage{makecell}
\usepackage{hyperref, cleveref}
\usepackage{pifont}
\usepackage[inkscapelatex=false]{svg}
\usepackage{caption}
\usepackage{wrapfig}
\DeclareCaptionLabelSeparator{pipe}{ | }
\usepackage{tcolorbox}
\newtcolorbox{coloredquote}[1][]{
    colback=green!5!white,  
    colframe=green!70!black, 
    boxrule=2pt,
    arc=7pt,
    left=6pt,
    right=6pt,
    top=4pt,
    bottom=4pt,
    title=#1
}

\setlist[itemize]{leftmargin=*}
\setlist[enumerate]{leftmargin=*}
\setlist[description]{leftmargin=*}

\title{ACE: Pluggable Adaptive Context Elasticizer across Agents}

\author{
Ning Liao$^{1*}$, Zihao Long$^{1,2*}$, Xiaoxing Wang$^{2}$, Xue Yang$^{2}$, Yaoming Wang$^1$, Ziyuan Zhuang$^1$,\\ \textbf{Xunliang Cai}$^1$, \textbf{Rongxiang Weng}$^{1\dagger}$, \textbf{Junchi Yan}$^{2\dagger}$ \\
    $^{1}$ Meituan, $^{2}$ Shanghai Jiao Tong University \\
    $^{*}$ Equal Contributions, $^{\dagger}$ Corresponding Authors \\
}

\begin{document}
\maketitle
\setcounter{footnote}{0}

\begin{abstract}
The increasing complexity of agentic tasks has led to rapidly growing trajectory lengths, which poses significant challenges for large language model (LLM) based agents with fixed context windows. Existing context management techniques, such as truncation and summarization, suffer from inherent inflexibility and irreversibility: once information is discarded or compressed, it cannot be recovered even when it becomes critically relevant in later decision steps. To address these limitations, we propose the Adaptive Context Elasticizer (ACE), a plug‑and‑play module that elastically orchestrates historical step information into the agent’s context at each decision step. ACE maintains a lossless message maintenance layer that stores both raw messages and compressed abstractions for each historical step, while a context orchestration layer adaptively assigns each step an elastic type as raw, abstract, or drop, at every decision step based on the current task state. This reversible design ensures that the main LLM always receives a compact yet information‑rich context. We adapt ACE to four diverse agent frameworks, including ReAct, DeepAgent, WebThinker, and MiroFlow, without training or architectural modifications. Experiments show that ACE consistently outperforms truncation and summarization baselines, and brings consistent performance gains across all four agent frameworks. 
\end{abstract}

\section{Introduction}
\label{sec:intro}
The practical success of large language models (LLMs)~\citep{team2025longcat, zeng2026glm, team2025kimi, chen2026minimax} in real-world scenarios has been largely enabled by agents~\citep{yao2023reactsynergizingreasoningacting, shinn2023reflexionlanguageagentsverbal, Wang_2024, wang2026autoagent}. In the classic ReAct~\citep{yao2023reactsynergizingreasoningacting} paradigm, LLMs usually act as the central brain that repeatedly reasons about the current task state, calls external tools, and integrates tool responses. The generated information at each step is appended verbatim to the trajectory. As task complexity grows, trajectory length increases linearly~\citep{schick2023toolformerlanguagemodelsteach, nakano2022webgptbrowserassistedquestionansweringhuman, jin2025searchr1trainingllmsreason, li2025websailornavigatingsuperhumanreasoning}. This poses major challenges for models with a fixed window size, leading to information density dropping, attention dispersing, and reasoning quality degrading. Therefore, agentic context management has become indispensable for solving complex and long-horizon tasks.

A range of techniques\citep{zhang2024surveymemorymechanismlarge, hu2026memoryageaiagents, du2026memoryautonomousllmagentsmechanisms} has been proposed to address the long-context bottleneck. The simplest, truncation~\citep{deepseekai2025deepseekv32pushingfrontieropen, liang2026genericagenttokenefficientselfevolvingllm,miromindteam2026mirothinker17h1heavyduty}, retains the most recent content while discarding the more distant historical context using a sliding window. This achieves zero computational cost at the expense of semantic integrity. Unlike truncation, compression does not discard content blindly but rather rewrites raw history into shorter forms. Existing work in this vein adopts three distinct designs. Threshold-triggered compression rewrites the history when a length budget is exceeded or a subtask is completed~\citep{kang2025aconoptimizingcontextcompression, wu2026resumunlockinglonghorizonsearch, sun2025scalinglonghorizonllmagent,qian2026memobrainexecutivememoryagentic}. Recurrent compression maintains a fixed-size memory state that is updated via learned mechanisms at every timestep~\citep{zhou2025mem1learningsynergizememory, chen2026iterresearchrethinkinglonghorizonagents, yu2025memagentreshapinglongcontextllm}. Agent-controlled compression places the compression decision in the hands of the LLM itself, exposing context management operations as actions or tools it can invoke~\citep{zhang2026memoryactionautonomouscontext, ye2025agentfoldlonghorizonwebagents,yao2026arcactivereflectiondrivencontext}.

These methods suffer from common flaws: inflexibility and irreversibility. In agent workflows, the relevance of a given historical step to the current decision evolves flexibly over time. Some steps are critical, offering unique details required at the present moment. Others are redundant, as their content is fully contained or overwritten by later steps. The remainder are simply dispensable. However, truncation and compression treat each historical step equally, ignoring the varying importance of different information, and foreclose the possibility of adaptation. Once content is dropped by truncation or abstracted by compression, the original information cannot be recovered. This becomes problematic when a previously discarded message turns out to be critical, or when a compressed summary fails to preserve a necessary detail.
 
The above analysis highlights a key requirement: at each decision step, the main LLM that acts as the agent’s primary reasoning module should receive high information density, achieved by adaptively rendering each historical message in its optimal form. Accordingly, we propose the \textbf{Adaptive Context Elasticizer (ACE)}.

\begin{figure}[t!]
  \centering
  \includegraphics[width=0.9\linewidth]{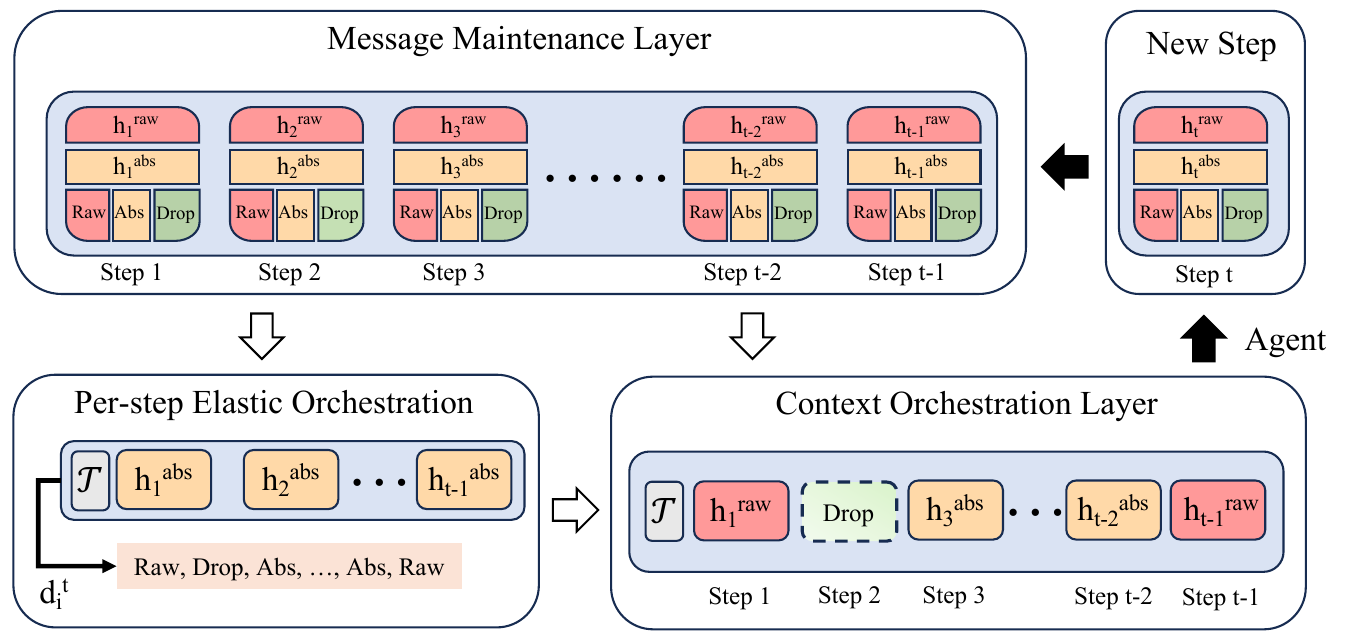}
  \caption{Overview of the proposed Adaptive Context Elasticizer (ACE). ACE comprises a message maintenance layer and a context orchestration layer. At each decision step, ACE performs per‑step elastic context orchestration by feeding historical abstractions as input, thereby constructing the actual context for the current step. Upon completion of the current action, the newly generated raw messages and compressed abstractions are fed back into the message maintenance layer for lossless storage.}
  \label{fig:ace_overview}
\end{figure}

As shown in Fig.~\ref{fig:ace_overview}, ACE distinguishes two layers: a message maintenance layer, which maintains the raw messages and compressed abstraction of each historical step, and a context orchestration layer, which produces the context that the main LLM actually receives. Before each step, ACE inspects the current task state and assigns one of three decisions, i.e., raw, abstract, or drop, to messages at each step in the message maintenance layer. These states collectively determine the composition of the context orchestration layer at the current step, ensuring the main LLM of the agents receives the most informative view of the trajectory for the decision at hand. Moreover, the presentation format of information from all historical steps is not fixed and can be adjusted on demand in any subsequent step. Given that the message maintenance layer remains immutable, an abstraction that is chosen for use in one step can be reversibly expanded to the full original information in another step. Similarly, information that is dropped in one step can be reversibly reassembled into the context in any subsequent step upon demand. This elastic context orchestration adapts synchronously with the agent's evolving focus of attention as the task progresses.

By this design, the inflexibility issue could be resolved as ACE is elastic, with the context orchestration layer rebuilt freshly based on the re-evaluated set of states at each step; and the irreversibility issue could be addressed as ACE is lossless, with the message maintenance layer never overwritten by any state assignment. To facilitate simple and flexible integration of ACE into various agent frameworks without substantially altering their native action spaces or message flow, we cast ACE as an external wrapper that encloses the reasoning loop of the original agent framework, thereby providing a more informative context at each step.

Experimentally, based on the classic ReAct agent framework~\citep{yao2023reactsynergizingreasoningacting}, we compare the proposed ACE against truncation and compression methods, showing that ACE can manage the context more effectively than the baseline methods. Moreover, as a training‑free, plug‑and‑play module, ACE is adapted to three other distinct agent frameworks: DeepAgent~\citep{li2026deepagentgeneralreasoningagent}, WebThinker~\citep{li2025webthinkerempoweringlargereasoning}, and MiroFlow~\citep{su2026miroflowhighperformancerobustopensource}. Across various benchmark evaluations, ACE consistently improves the performance over the original frameworks.

We summarize the contributions as follows:
\begin{enumerate}
  \item \textbf{Decoupling Actual Context from Raw Messages.} We devise a maintenance layer for raw messages and compressed abstractions, and an orchestration layer for actual context. This lossless design enables reversible expansion of abstracts and recovery of dropped messages in later steps.
  
  \item \textbf{Per-step Elastic Context Orchestration.} We reorganize every historical message into \texttt{Raw}, \texttt{Abstract}, or \texttt{Drop} based on the current task state at each step, then rebuild the context orchestration layer to minimize redundancy and maximize information density without altering storage.
  
  \item \textbf{Training-free Plug-and-Play Integration.} We implement ACE as an external wrapper around an existing agent's reasoning loop without modifying action space, message flow, or training, and adapt it to four diverse frameworks: ReAct, WebThinker, DeepAgent, and MiroFlow.
  
  \item \textbf{Empirical Analysis and Ablation.} ACE outperforms truncation and compression on ReAct, and improves performance over the original frameworks when adapted to WebThinker, DeepAgent, and MiroFlow. Further ablation studies confirm the efficacy of the mechanism in the proposed ACE.
\end{enumerate}

\section{Related Work}
\label{sec:related}
\subsection{Agentic Context Management}
Existing approaches to agentic context management follow four broad designs. \textit{Truncation}~\citep{deepseekai2025deepseekv32pushingfrontieropen,liang2026genericagenttokenefficientselfevolvingllm,miromindteam2026mirothinker17h1heavyduty} methods discard trajectory content by position or length, which is entirely content‑agnostic. \textit{Threshold‑triggered compression} methods, such as ACON~\citep{kang2025aconoptimizingcontextcompression} and ReSum~\citep{wu2026resumunlockinglonghorizonsearch}, invoke a compression LLM to summarize the accumulated history when the context length surpasses a budget. The resulting summary then replaces the original trajectory. \textit{Recurrent compression} methods, exemplified by MEM1~\citep{zhou2025mem1learningsynergizememory}, MemAgent~\citep{yu2025memagentreshapinglongcontextllm}, and IterResearch~\citep{chen2026iterresearchrethinkinglonghorizonagents}, train the primary reasoning LLM end‑to‑end to emit a fixed‑size summary state at each step. \textit{Agent‑controlled compression} methods expose context management to the main LLM as explicit actions or tools. For example, AgentFold~\citep{ye2025agentfoldlonghorizonwebagents} introduces a fold action that retrospectively condenses individual steps or consolidates multi‑step sub‑tasks, whereas Memory‑as‑Action~\citep{zhang2026memoryactionautonomouscontext} formulates precise compression operations as actions governed by a trained policy. Across all four families, every compression decision is final and cannot be revised in later steps. Moreover, recurrent and agent‑controlled methods modify the underlying framework, extend the action space or tool set, and require additional training. These two limitations motivate ACE, which retains the full trajectory in a separate storage layer so that every assignment remains revisable and flexible, and operates as an external wrapper so that no change to the original agent frameworks or any further training is needed.

\subsection{Agentic Foundation Models}
To elevate agent performance, a complementary line of work is to enhance the intrinsic agentic abilities of foundation models~\citep{team2025longcat, zeng2026glm, team2025kimi, chen2026minimax}. The key design axis is context scaling, which extends the maximum context length that the model can ingest, for instance via sparse or recurrent attention~\citep{beltagy2020longformerlongdocumenttransformer, zaheer2021bigbirdtransformerslonger, yuan2025nativesparseattentionhardwarealigned, deepseekai2025deepseekv32pushingfrontieropen} or via positional interpolation~\citep{chen2023extendingcontextwindowlarge, peng2026yarnefficientcontextwindow}. These advances increase the maximum trajectory length an agent can handle before hitting the window limit, yet they fail to resolve a deeper problem: even within the supported window, models attend non‑uniformly to long inputs and their performance drops well before the nominal bound~\citep{laban2025llmslostmultiturnconversation, liu2023lostmiddlelanguagemodels, bai2024longbenchbilingualmultitaskbenchmark}. ACE takes an orthogonal approach: it stays within the LLM's existing context limit but focuses on raising the information density of the presented context, rather than expanding the window size.

\section{Method}
\label{sec:method}
In this section, we present the Adaptive Context Elasticizer (ACE) in detail. We first propose to decouple the construction of the actual decision-making context from the step-wise raw messages within an agent framework in Sec.~\ref{sec:problem}, in which we devise the two-layer framework for message maintenance and context orchestration in ACE. Subsequently, we describe ACE's paradigm of elastically orchestrating historical messages into context at each step in Sec.~\ref{sec:elastic_mechanism}. Finally, we demonstrate how to adapt ACE to existing agent frameworks in a plug-and-play manner in Sec.~\ref{sec:integration}.

\subsection{Decoupling Actual Context from Raw Messages}
\label{sec:problem}
Given a task $\mathcal{T}$ to be resolved by the agent, at each step $t$, it outputs the reasoning $r_t$ on the current state, produces an action $a_t$ for tool invocation, and receives the observation $o_t$. Then, the newly generated information at step $t$ is packed as $h_t^{\text{raw}} = (r_t, a_t, o_t)$ and will be appended into the trajectory. We denote $\mathcal{H}_t = \{h_1^{\text{raw}}, \dots, h_{t-1}^{\text{raw}}\}$ as the raw messages accumulated before step $t$, under the ReAct paradigm, the input of the agent at step $t$ is $\mathcal{H}_t$. With the accumulation of raw stepwise information over a trajectory, the growing redundancy and surging volume pose significant challenges for the main LLMs of agents in reasoning and making decisions effectively. Existing agent frameworks often manage $\mathcal{H}_t$ in place through linear concatenation, threshold-triggered compression, or recurrent rewriting. In such designs, any decision becomes irrevocable: once a step is truncated or compressed, the original information is permanently lost.

To this end, we propose to decouple the raw messages $\mathcal{H}_t$ from the context $\mathcal{C}_t$ that the agents actually receive at step $t$. To effectuate this decoupling design, we devise a two-layer framework in ACE, including a message maintenance layer for storing raw messages and compressed abstractions, and a context orchestration layer that provides the agent with the actual context.

\textbf{Message Maintenance Layer.} We denote this layer by $\mathcal{M}$, and it is formulated as:
\begin{equation}
\begin{aligned}
\label{eq:mm_layer}
\mathcal{M} = \bigcup_{i=1}^{t}(h_i^{\text{raw}}, h_i^{\text{abs}}),
\end{aligned}
\end{equation}
where $\bigcup$ is the union operator. Upon completion of a step $i$, the raw message $h_i^{\text{raw}}$ generated from that step, including reasoning, actions, and observations, is recorded in $\mathcal{M}$. In addition, we also preserve a compressed abstraction $h_i^{\text{abs}}$ summarized by an auxiliary LLM for each step. The abstraction leads with a conclusive result; preserves task‑critical facts such as numbers, URLs, dates, IDs, formulas, quotations, and tool schemas; renders tool call failures explicit by stating the attempted action, the cause of failure, and the measures to avoid; and retains explicit follow‑up signals. Furthermore, once the raw information and compressed abstraction $(h_i^{\text{raw}}, h_i^{\text{abs}})$ of this step are recorded, they are never modified, thereby preserving the completeness and traceability of the information from each historical step.

\textbf{Context Orchestration Layer.} Unlike existing agent frameworks that feed the main LLM with either a lengthy concatenation of raw messages, or irreversibly compressed or truncated content at each step, we adaptively tailor the context to the evolving task state. Specifically, the actual context $\mathcal{C}_t$ consists of three segments:

The task description $\mathcal{T}$ is placed at the start of the context and remains free from any summarization or truncation. This guarantees that the agent's reasoning and action execution stay aligned with the task.

As for the historical steps ranging from $1$ to $t-2$, we adaptively incorporate their information into the context in different forms, conditioned on the current task state. For historical steps that contain the specific details and essential justifications required for the current decision step, their raw messages are selected. For steps that provide transitional content or high‑level overviews relevant to the current decision step, their compressed abstraction is selected. For steps that are irrelevant to the current decision step, or whose information has been modified or superseded by subsequent steps, they are omitted from the current context entirely.

The final part of the context is fixed as the complete information of the last step, enabling the current decision step to have a full understanding and assessment of the latest state of task execution.

This design reduces the redundancy that occupies the context window, leading to more effective reasoning and action execution. Formally, the actual context $\mathcal{C}_t$ is mainly orchestrated through the function $\phi$:
\begin{equation}
\begin{aligned}
\label{eq:col}
\mathcal{C}_t =\; \text{Concat}\;(\mathcal{T}, \phi\left(h_1^{\text{raw}},h_1^{\text{abs}},\,d_1^{t}\right), ..., \phi\left(h_{t-2}^{\text{raw}},h_{t-2}^{\text{abs}},d_{t-2}^{t}\right), h_{t-1}^{\text{raw}})
\end{aligned}
\end{equation}
where $d_i^{t} \in \{\text{Raw},\,\text{Abs},\,\text{Drop}\}$ denotes the decision made at step $t$ regarding how the information of historical step $i \in [1, t-2]$ is orchestrated in the actual context $\mathcal{C}_t$ input to the agent. The function $\phi$ then applies the per-step selection according to the decision $d_i^{t}$, which outputs the raw data $h_i^{\text{raw}}$ when $d_i^{t}$ is \text{Raw}, or the compressed abstraction $h_i^{\text{abs}}$ when $d_i^{t}$ is \text{Abs}, or nothing when $d_i^{t}$ is \text{Drop}. 
 
\subsection{Per-step Elastic Context Orchestration}
\label{sec:elastic_mechanism}
Building upon the complete preservation of historical information by the message maintenance layer and the composition of the actual context by the context orchestration layer, we present the detailed mechanism of ACE's per-step elastic orchestration from historical information to the actual context.

The core is the step-wise decision $d_i^{t} \in \{\text{Raw},\,\text{Abs},\,\text{Drop}\}$ predicted by the elasticizer $\mathcal{E}$, which is instantiated as an LLM. Its input is the task description $\mathcal{T}$ together with the abstractions $\{h_i^{\text{abs}}\}_{i=1}^{t-1}$ drawn from the message maintenance layer $\mathcal{M}$, formulated as:
\begin{equation}
\begin{aligned}
\label{eq:els}
d_i^{t} = \mathcal{E}(\mathcal{T},h_1^{\text{abs}},...,h_{t-1}^{\text{abs}}).
\end{aligned}
\end{equation}

Through this design, the elasticizer gains full visibility into all historical information while keeping its input length within the sum of per‑step abstraction lengths. As a result, the elasticizer can reliably predict orchestration decisions within the effective context window. It is worth noting that the elasticizer's decisions are advisory for the interior portion of the context, i.e., steps ranging from $1$ to $t-2$, but have no effect on the fixed boundary positions. Consequently, the most recent step $t-1$ is always included in its raw messages as formulated in Eq.~\ref{eq:col}, irrespective of the elasticizer's output.

Due to the decoupling of historical information from the actual context and the per‑step orchestration of the context, ACE provides reversibility and flexibility. A historical step compressed at one decision step can be restored to its raw form in subsequent decision steps when necessary, and a discarded historical step can be retrieved as needed. Hence, the non‑boundary part of the context becomes genuinely elastic, modulated by the agent's dynamically evolving states.
 
\subsection{Training-free Plug-and-Play Integration}
\label{sec:integration}
Since ACE runs completely on top of its own two‑layer architecture without altering the agent's state, action space, or control flow, its adaptation to existing agent frameworks requires no training and is plug‑and‑play. Specifically, after the agent completes its reasoning, action execution, and observation acquisition at each step, the raw messages and generated corresponding abstraction from each step are stored in the message maintenance layer. Then, before the agent begins its reasoning at each step, the actual context is organized by the context orchestration layer. This lightweight interface enables the instantiation of ACE on four agent frameworks with distinct structures: ReAct~\citep{yao2023reactsynergizingreasoningacting}, WebThinker~\citep{li2025webthinkerempoweringlargereasoning}, DeepAgent~\citep{li2026deepagentgeneralreasoningagent}, and MiroFlow~\citep{su2026miroflowhighperformancerobustopensource}. Collectively, these frameworks span single‑agent, main‑auxiliary, and multi‑agent architectural topologies. 


\section{Experiments}
\label{sec:experiments}
We begin by detailing the experimental setup in Sec.~\ref{sec:exp_setup}. Then, Sec.~\ref{sec:exp_baselines} presents a comparison between our proposed ACE method and baseline methods based on the ReAct framework. In Sec.~\ref{sec:exp_across}, we validate the plug‑and‑play adaptability of ACE across multiple agent frameworks beyond ReAct by applying it to DeepAgent~\citep{li2026deepagentgeneralreasoningagent}, WebThinker~\citep{li2025webthinkerempoweringlargereasoning}, and MiroFlow~\citep{su2026miroflowhighperformancerobustopensource}, and demonstrate the resulting performance improvements. Furthermore, we present a series of analytical experiments on the per-step input token length in Sec.~\ref{sec:ctx_len}, elastic type evolution of historical steps in Sec.~\ref{sec:ctx_type}, ablation on the composition of elastic types in Sec.~\ref{sec:elas_type}, and ablation on whether to keep the last step as raw in Sec.~\ref{sec:last_step}.

\subsection{Experimental Setup}
\label{sec:exp_setup}
In our experiments, we choose GPT-4.1 and Gemini-3.1-flash-lite-preview as the main LLMs of the agent frameworks for generality. As for the proposed ACE, the elasticizer $\mathcal{E}$ is set to be the same as the main LLM, and the auxiliary LLM used for generating the compressed abstraction is uniformly set to GPT-4o. Regarding the tools, to support the web search, we use the Google Serper API. The page browsing is implemented using the Jina Reader API. Besides, for comprehending the images, the multimodal understanding model is also implemented using GPT-4o. Considering the benchmarks, in the comparison with baseline methods in Sec.~\ref{sec:exp_baselines}, we select GAIA~\citep{mialon2023gaiabenchmarkgeneralai}, HLE~\citep{phan2025humanity}, and WebShop~\citep{yao2023webshopscalablerealworldweb} for evaluation. In experiments adapting ACE to more agent frameworks in Sec.~\ref{sec:exp_across}, we select the benchmarks adopted by the original agents. In detail, GAIA, HLE, and WebShop are selected for evaluation in DeepAgent;  GAIA, WebWalkerQA~\citep{wu2025webwalkerbenchmarkingllmsweb}, and HLE are selected for evaluation in WebThinker; GAIA, xBench-DS~\citep{chen2025xbenchtrackingagentsproductivity}, and BrowseComp-ZH~\citep{zhou2025browsecompzhbenchmarkingwebbrowsing} are selected for evaluation in MiroFlow. All experiments are implemented from scratch by our setups.

\subsection{Comparison with Context Management Baselines}
\label{sec:exp_baselines}
Based on the classical ReAct paradigm, we compare our proposed ACE method with existing context management baselines, including no context management, truncation‑based management, and summarization‑based management. Specifically, no context management refers to the original design of ReAct. As for the implementation of the truncation method, we retain the raw messages of the latest 5 steps, while others are discarded. Besides, we implement the summarization method such that the raw messages of the latest 5 steps are retained, while others are summarized. We set the main LLMs as GPT-4.1 and Gemini-3.1-flash-lite-preview for generality, and limit the maximum action steps to 50. The experimental results are presented in Table~\ref{tab:bsl_comp}.
\begin{table}[t!]
\centering
\caption{The comparison between the proposed ACE and context management baseline methods, including no context management, truncation and summarization, based on the ReAct Framework.}
\label{tab:bsl_comp}
\begin{tabular}{p{4.2cm}<{\centering} | p{0.8cm}<{\centering} p{0.8cm}<{\centering} p{0.8cm}<{\centering} p{0.8cm}<{\centering} | p{0.8cm}<{\centering} p{0.8cm}<{\centering} p{0.8cm}<{\centering} | p{0.9cm}<{\centering} p{0.9cm}<{\centering}} 
\toprule
\multirow{2}{*}{\textbf{Benchmarks}} & \multicolumn{4}{c|}{\textbf{GAIA}} & \multicolumn{3}{c|}{\textbf{HLE}} & \multicolumn{2}{c}{\textbf{WebShop}} \\
& Text & MM & File & All & Text & MM & All & Success & Score \\
\midrule
\multicolumn{10}{c}{\textit{Main LLM: GPT-4.1}} \\
\midrule
ReAct                       & 46.7 & 33.3 & 21.1 & 38.8 & 22.5 & 21.2 & 22.2 & \textbf{22.8} & 55.1 \\
ReAct + Truncation          & 43.7 & 33.3 & 21.1 & 37.0 & 23.5 & 23.0 & 23.4 & 21.2 & 53.4 \\
ReAct + Summarization       & 49.5 & 29.2 & 21.1 & 40.0 & 23.0 & \textbf{23.0} & 23.0 & 22.0 & 55.1 \\
\textbf{ReAct + ACE (Ours)} & \textbf{51.5} & \textbf{33.3} & \textbf{23.7} & \textbf{42.4} & \textbf{24.8} & 22.1 & \textbf{24.2} & 20.8 & \textbf{56.7} \\
\midrule
\multicolumn{10}{c}{\textit{Main LLM: Gemini-3.1-flash-lite-preview}} \\
\midrule
ReAct                      & 51.5 & 41.7 & \textbf{34.2} & 46.1 & \textbf{29.7} & 26.5 & 29.0 & 34.8 & 62.6 \\
ReAct + Truncation         & 51.5 & 29.2 & 28.9 & 43.0 & 29.2 & 28.3 & 29.0 & 35.6 & 62.0 \\
ReAct + Summarization      & 46.6 & 29.2 & 23.7 & 38.8 & 25.8 & 26.5 & 26.0 & \textbf{36.0} & 62.7 \\
\textbf{ReAct + ACE (Ours)}         & \textbf{62.1} & \textbf{50.0} & 28.9 & \textbf{52.7} & 29.5 & \textbf{29.2} & \textbf{29.4} & 35.2 & \textbf{63.4} \\
\bottomrule
\end{tabular}
\end{table}
\begin{table}[t!]
\centering
\caption{The performance of adapting the proposed ACE to the DeepAgent.}
\label{tab:ACE_DeepAgent}
\begin{tabular}{p{4.2cm}<{\centering} | p{0.8cm}<{\centering} p{0.8cm}<{\centering} p{0.8cm}<{\centering} p{0.8cm}<{\centering} | p{0.8cm}<{\centering} p{0.8cm}<{\centering} p{0.8cm}<{\centering} | p{0.9cm}<{\centering} p{0.9cm}<{\centering}} 
\toprule
\multirow{2}{*}{\textbf{Benchmarks}} & \multicolumn{4}{c|}{\textbf{GAIA}} & \multicolumn{3}{c|}{\textbf{HLE}} & \multicolumn{2}{c}{\textbf{WebShop}} \\
& Text & MM & File & All & Text & MM & All & Success & Score \\
\midrule
\multicolumn{10}{c}{\textit{Main LLM: GPT-4.1}} \\
\midrule
DeepAgent                       & 45.6 & 41.7 & \textbf{23.7} & 40.0 & 23.3 & 21.2 & 22.8 & 20.0 & 52.2 \\
\textbf{DeepAgent + ACE (Ours)}          & \textbf{49.5} & \textbf{41.7} & 21.1 & \textbf{41.8} & \textbf{23.5} & \textbf{21.2} & \textbf{23.0} & \textbf{20.8} & \textbf{57.3} \\
\midrule
\multicolumn{10}{c}{\textit{Main LLM: Gemini-3.1-flash-lite-preview}} \\
\midrule
DeepAgent                       & 55.3 & 41.7 & \textbf{26.3} & 46.7 & 27.6 & 23.0 & 26.6 & \textbf{36.4} & 61.2 \\
\textbf{DeepAgent + ACE (Ours)}         & \textbf{61.2} & \textbf{54.2} & 23.7 & \textbf{51.5} & \textbf{28.6} & \textbf{26.5} & \textbf{28.2} & 34.8 & \textbf{63.8} \\
\bottomrule
\end{tabular}
\end{table}

As reported in Table~\ref{tab:bsl_comp}, ACE achieves the best overall performance among all compared methods under both main LLMs across the three benchmarks. For example, on GAIA, ACE improves the overall score from 38.8 to 42.4 when using GPT‑4.1, and from 46.1 to 52.7 when using Gemini‑3.1‑flash‑lite‑preview, outperforming the baselines. The baseline behaviors corroborate our motivation. Truncation consistently degrades performance on GAIA, whose tasks may require evidence from earlier steps that a sliding window irreversibly discards; yet it improves performance on HLE relative to vanilla ReAct, indicating that the redundancy in raw trajectories indeed dilutes information density. Summarization partially mitigates information loss on GAIA with GPT‑4.1 but fails when we use Gemini‑3.1‑flash‑lite‑preview, suggesting that a one‑shot, irreversible summary may omit details whose importance only becomes apparent later. By reassessing the optimal form of every historical step at each turn and losslessly restoring or dropping content on demand, the proposed ACE sustains high information density in the context, thereby enabling more effective reasoning compared with baselines.

\subsection{Adapting ACE as a Plug-in to Multiple Agents}
\label{sec:exp_across}
ACE is designed to be a plug‑and‑play context manager that can be seamlessly integrated into different agent frameworks. To verify its generality, this section adapts ACE to three frameworks beyond ReAct, including DeepAgent~\citep{li2026deepagentgeneralreasoningagent}, WebThinker~\citep{li2025webthinkerempoweringlargereasoning}, and MiroFlow~\citep{su2026miroflowhighperformancerobustopensource}. The performance is evaluated on the same benchmarks as those originally reported for each.
\begin{table}[t!]
\centering
\caption{The performance of adapting the proposed ACE to the WebThinker.}
\label{tab:ACE_WebThinker}
\begin{tabular}{p{4.2cm}<{\centering} | p{0.8cm}<{\centering} p{0.8cm}<{\centering} p{0.8cm}<{\centering} p{0.8cm}<{\centering} | p{0.8cm}<{\centering} p{0.8cm}<{\centering} p{0.8cm}<{\centering} p{0.8cm}<{\centering} | p{0.9cm}<{\centering}} 
\toprule
\multirow{2}{*}{\textbf{Benchmarks}} & \multicolumn{4}{c|}{\textbf{GAIA}} & \multicolumn{4}{c|}{\textbf{WebWalkerQA}} & \multicolumn{1}{c}{\textbf{HLE}} \\
& L1 & L2 & L3 & AVG & Easy & Med. & Hard & AVG & AVG \\
\midrule
\multicolumn{10}{c}{\textit{Main LLM: GPT-4.1}} \\
\midrule
WebThinker                       & 46.2 & 26.9 & 8.3 & 32.0 & 25.6 & 22.5 & \textbf{20.0} & 22.4 & 6.8 \\
\textbf{WebThinker + ACE (Ours)}          & \textbf{53.8} & \textbf{42.3} & \textbf{8.3} & \textbf{42.7} & \textbf{25.6} & \textbf{28.2} & 19.6 & \textbf{24.6} & \textbf{9.8} \\
\midrule
\multicolumn{10}{c}{\textit{Main LLM: Gemini-3.1-flash-lite-preview}} \\
\midrule
WebThinker                       & 38.5 & 30.8 & 16.7 & 32.0 & \textbf{28.8} & 20.0 & 17.1 & 21.0 & 9.0 \\
\textbf{WebThinker + ACE (Ours)}          & \textbf{48.7} & \textbf{40.4} & \textbf{16.7} & \textbf{40.8} & 20.0 & \textbf{22.9} & \textbf{20.4} & \textbf{21.3} & \textbf{11.2} \\
\bottomrule
\end{tabular}
\end{table}
\vspace{-10pt}
\begin{table}[t!]
\centering
\caption{The performance of adapting the proposed ACE to the MiroFlow.}
\label{tab:ACE_MiroFlow}
\begin{tabular}{p{4.2cm}<{\centering} | p{0.8cm}<{\centering} p{0.8cm}<{\centering} | p{2.0cm}<{\centering} | p{3.0cm}<{\centering}} 
\toprule
\multirow{2}{*}{\textbf{Benchmarks}} & \multicolumn{2}{c|}{\textbf{GAIA Val}} & \multirow{2}{*}{\textbf{xBench-DS}} & \multirow{2}{*}{\textbf{BrowseComp-ZH}}  \\
& AVG & Text & & \\
\midrule
\multicolumn{5}{c}{\textit{Main LLM: GPT-4.1}} \\
\midrule
MiroFlow                       & 33.3 & 34.0 & 33.0 & 9.7 \\
\textbf{MiroFlow + ACE (Ours)}          & \textbf{43.6} & \textbf{43.7} & \textbf{37.0} & \textbf{12.5} \\
\midrule
\multicolumn{5}{c}{\textit{Main LLM: Gemini-3.1-flash-lite-preview}} \\
\midrule
MiroFlow                       & 64.2 & 56.3 & 57.0 & 21.5 \\
\textbf{MiroFlow + ACE (Ours)}          & \textbf{66.1} & \textbf{61.2} & \textbf{61.0} & \textbf{26.6} \\
\bottomrule
\end{tabular}
\end{table}

Table~\ref{tab:ACE_DeepAgent} reports the results on DeepAgent after its native memory fold mechanism is replaced with ACE. On GAIA, ACE raises the overall performance from 40.0 to 41.8 with GPT‑4.1 and from 46.7 to 51.5 with Gemini‑3.1‑flash‑lite‑preview, while also achieving consistent improvements on HLE and WebShop. These findings indicate that ACE successfully outperforms the original memory fold mechanism. As reported in Table~\ref{tab:ACE_WebThinker}, ACE brings larger gains when applied to the WebThinker. The overall performance of GAIA increases by at least 8 points, accompanied by a marked improvement on the L2 subset. This pattern is consistent with the view that extended trajectories are especially vulnerable to redundant information, which our method successfully alleviates. Beyond GAIA, ACE also yields performance improvements on WebWalkerQA and HLE. These findings suggest that for WebThinker, which does not incorporate any context management by default, the adoption of ACE leads to consistent performance gains. Furthermore, when ACE is adapted to the MiroFlow framework, consistent performance improvements are observed across two models and three benchmarks in Table~\ref{tab:ACE_MiroFlow}. Notably, on the GAIA benchmark with GPT‑4.1 as the main LLM, ACE achieves a substantial performance advantage of nearly 10 points. These results demonstrate that ACE effectively reduces redundant information in each decision‑making step, thereby enhancing the final performance.

Across all four agent frameworks, i.e., ReAct, DeepAgent, WebThinker, and MiroFlow, ACE consistently improves the performance with neither training nor modification to the native action spaces. This indicates that ACE is a genuinely plug‑and‑play context manager, its effectiveness being agnostic to the underlying agent architecture. The combination of generality, seamless integration, and consistent gains positions ACE as a readily deployable enhancement for existing and future agent frameworks.
 
\subsection{Analysis on the Per-Step Input Token Length}
\label{sec:ctx_len}
\begin{figure}[t!]
\begin{minipage}[b]{8.2cm}
  \centering
  \centerline{\includegraphics[width=8.2cm]{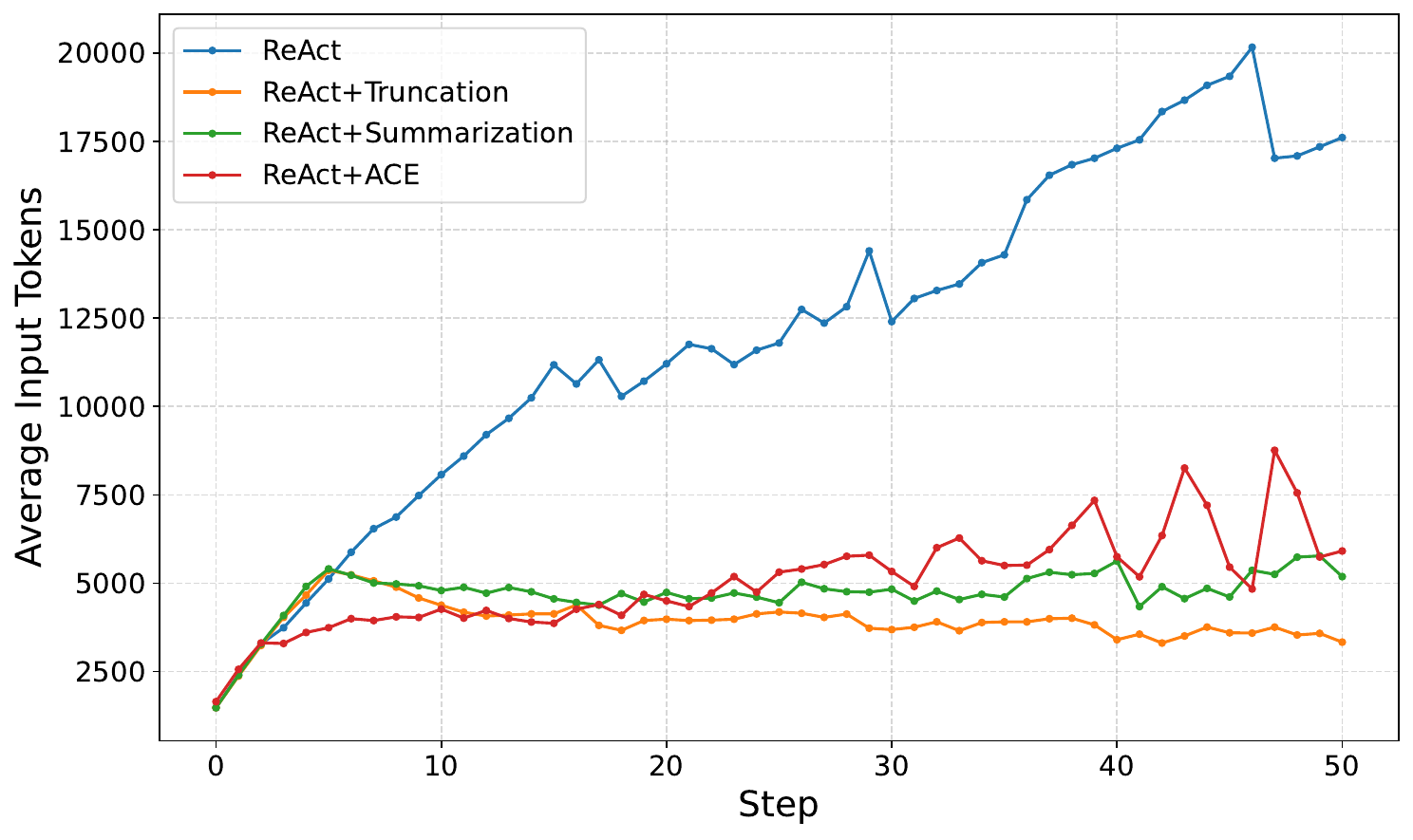}}
  \centerline{\footnotesize (a) Per‑step input token length (max action steps = 50). }
\end{minipage}
\begin{minipage}[b]{8.2cm}
  \centering
  \centerline{\includegraphics[width=8.2cm]{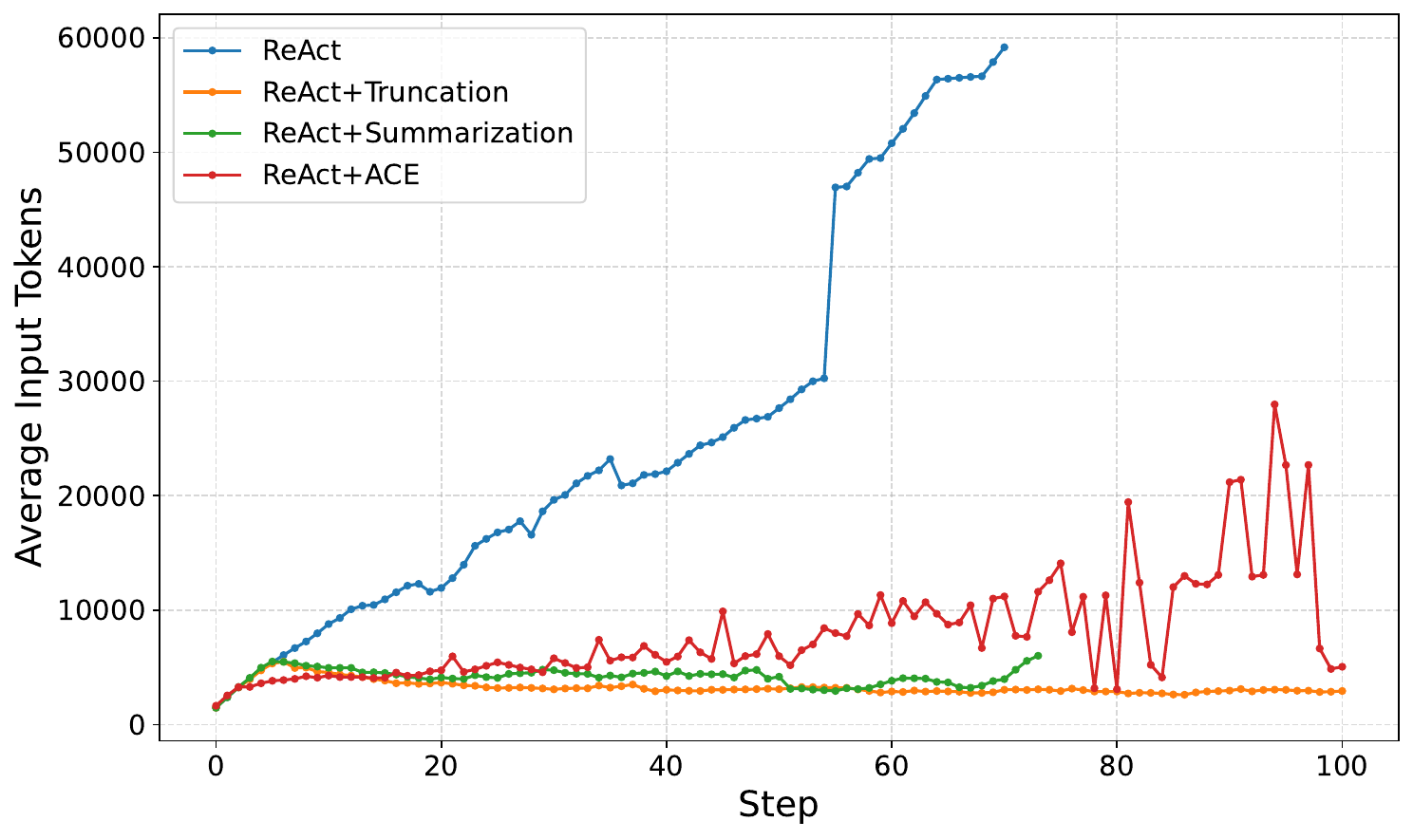}}
  \centerline{\footnotesize (b) Per‑step input token length (max action steps = 100). }
\end{minipage}
\caption{The analysis on the average per-step input token length using Gemini‑3.1‑flash‑lite‑preview as the main LLM on the GAIA benchmark by setting the max action steps as 50 and 100 across 4 settings, including ReAct, ReAct with truncation, ReAct with summarization, and ReAct with the proposed ACE.} 
\label{fig:ctx_growth}
\end{figure}

To further investigate how ACE influences the adaptive organization of context during task execution, we conduct a comparative analysis using Gemini‑3.1‑flash‑lite‑preview as the main LLM on the GAIA benchmark. Specifically, we compare the average per‑step input token length of the main LLM on all tasks across four settings, including ReAct, ReAct with truncation, ReAct with summarization, and ReAct with the proposed ACE, under maximum action step limits of 50 and 100. The results are presented in Fig.~\ref{fig:ctx_growth}.

Under both maximum action step constraints, vanilla ReAct without any context management exhibits a continuously increasing trend in per‑step input token length as the number of executed steps grows. This leads to challenges such as exceeding the LLM's context window and accumulating redundant information. For the two baseline methods, summarization and truncation both retain only the raw messages of the most recent five steps. Additionally, summarization includes a summary of information from steps beyond the most recent five, whereas truncation does not, resulting in the per‑step input token length of summarization being slightly higher than that of truncation. Consequently, both approaches maintain per‑step input token length within a relatively low range, effectively suppressing the linear growth of input tokens. However, as shown in Table~\ref{tab:bsl_comp}, suppressing the growth of the input context may incur negative performance impacts. For example, when Gemini‑3.1‑flash‑lite‑preview is used as the main LLM on GAIA, both ReAct with truncation and summarization achieve lower overall performance than vanilla ReAct. This indicates that these two context management approaches lack flexibility and inevitably cause irreversible loss of important information from distant steps, thereby compromising overall task performance.

As for the proposed ACE method, similar to the two baselines, it achieves a per‑step input token length significantly lower than that of ReAct. When we compare ACE with the two baselines in the early steps, ACE exhibits a slightly lower average input token length. This is because the baselines have not yet activated their management mechanisms at the early stage, thus retaining raw historical messages throughout, whereas ACE's elastic orchestration naturally yields a shorter input. As the task proceeds, ACE's average input token length overtakes that of the baselines and exhibits a gradually increasing yet fluctuating trend. This observation indicates that ACE does not rigidly constrain the input length within a fixed range. Instead, it adaptively and dynamically orchestrates the presentation form of each historical step for the current decision step according to the evolving task state, flexibly discarding redundant messages, retaining necessary information, and even reversibly recalling information that was discarded in previous decision steps. 

In summary, simply relying on truncation or summarization to reduce context pressure is not an optimal solution, as it may lead to information loss. The effectiveness of our proposed ACE method, which flexibly and elastically organizes the context according to the task state, is thus fully validated.

\subsection{Analysis on the Elastic Type Evolution of Historical Steps}
\label{sec:ctx_type}

\begin{figure}[t!]
\begin{minipage}[b]{8.2cm}
  \centering
  \centerline{\includegraphics[width=7.8cm]{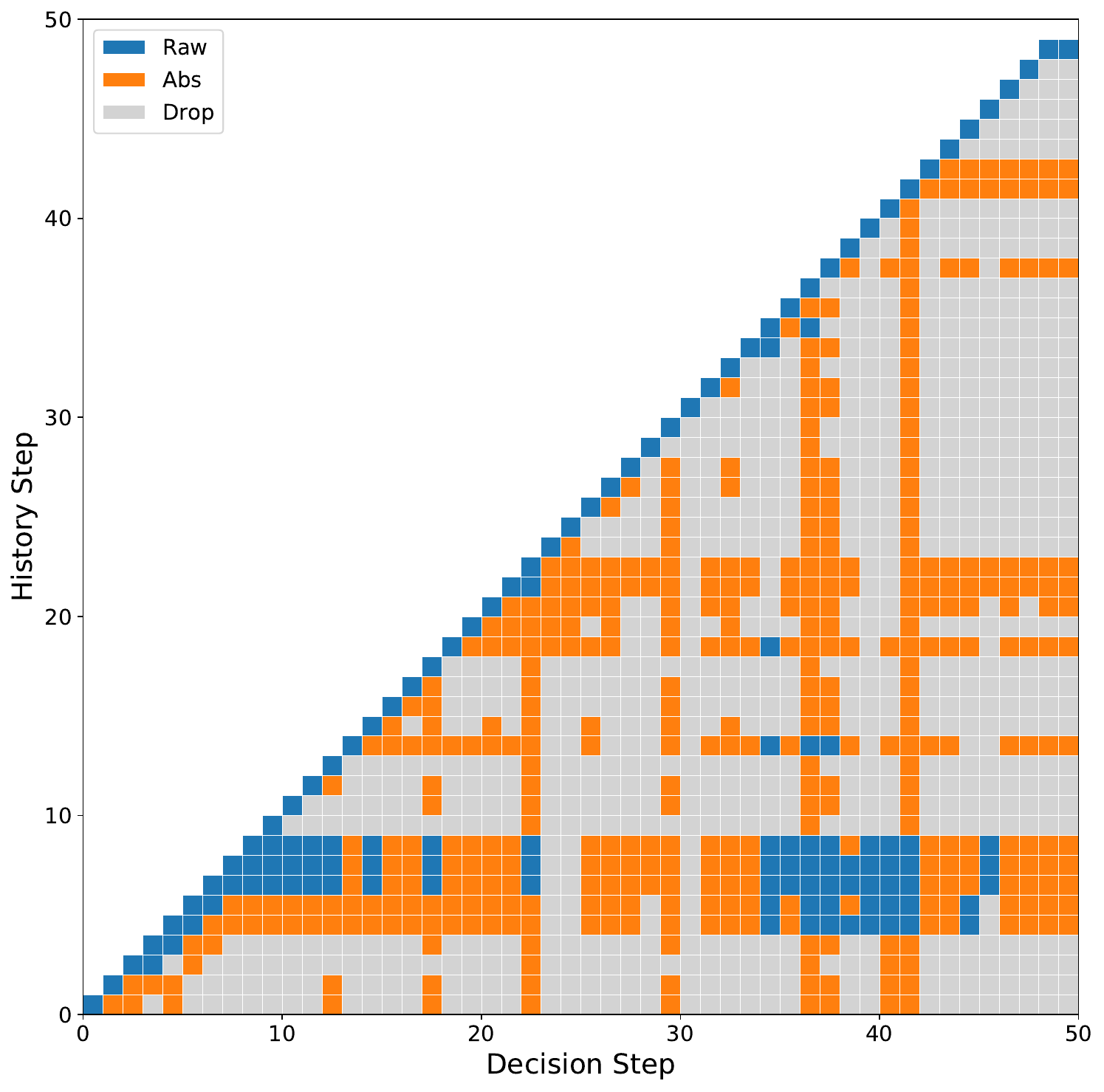}}
  \centerline{\footnotesize (a) Visualization of one task from GAIA. }
\end{minipage}
\begin{minipage}[b]{8.2cm}
  \centering
  \centerline{\includegraphics[width=7.8cm]{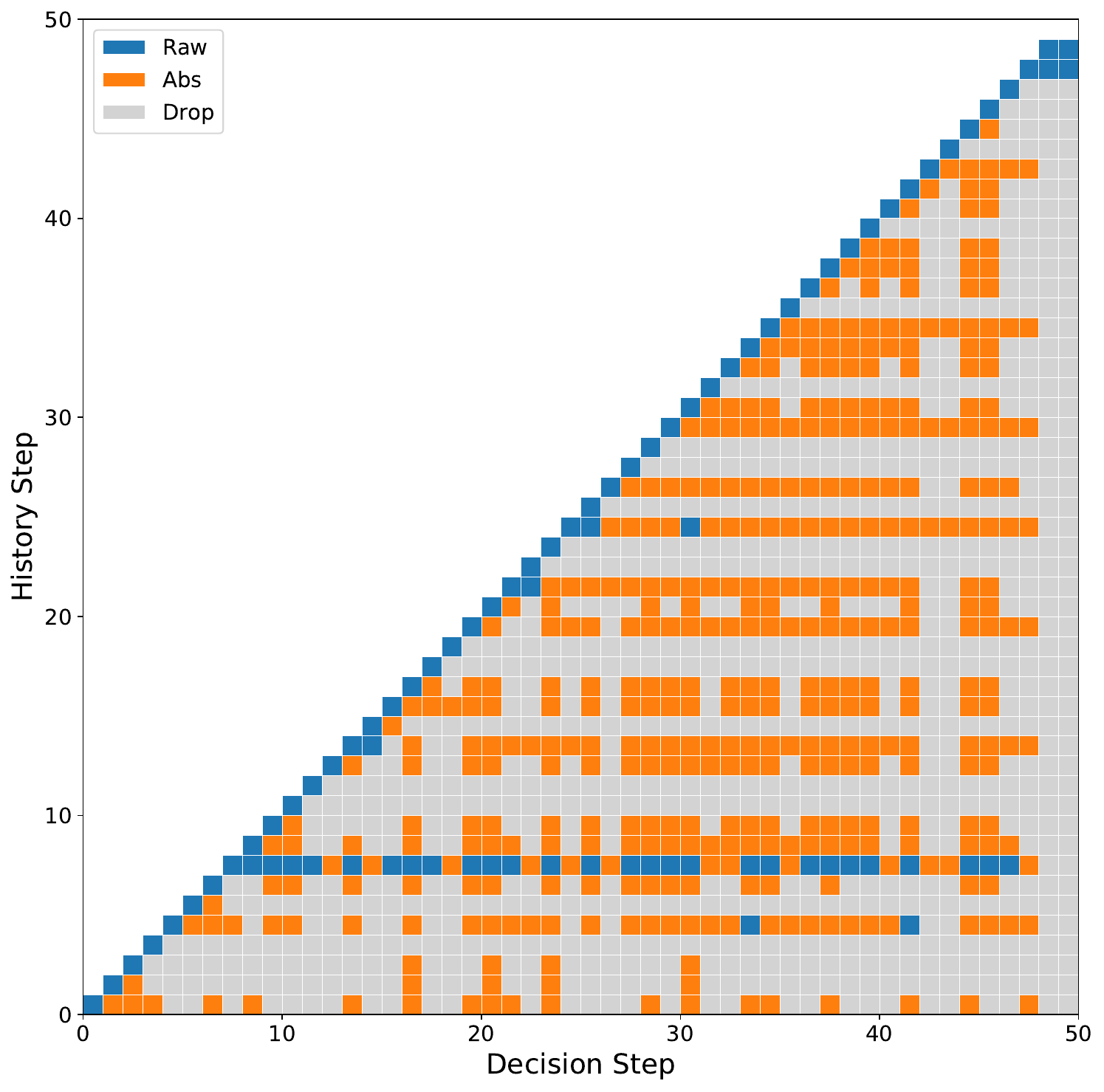}}
  \centerline{\footnotesize (b) Visualization of one task from HLE. }
\end{minipage}
\caption{Visualization of the elastic type evolution of historical steps under the setting where ACE is adapted to the ReAct framework with Gemini‑3.1‑flash‑lite‑preview as the main LLM, based on one task selected from GAIA and one from HLE, respectively.} 
\label{fig:ctx_type}
\end{figure}
Beyond the analysis of per-step input token length, observing and analyzing how the elastic type of each historical step evolves within the context of each decision-making step as the task progresses can further advance the understanding of the mechanism of ACE. Therefore, in the setting where Gemini‑3.1‑flash‑lite‑preview serves as the main LLM and ACE is adapted to the ReAct framework, we select one task from GAIA and one from HLE. Based on their trajectories, we visualize the evolution of the elastic type of historical steps over the decision steps, as shown in Fig.~\ref{fig:ctx_type}.

By design, the last step in the context is always retained as raw messages. Therefore, all diagonal blocks are marked as raw. Proceeding horizontally from left to right illustrates how the elastic type of the same historical step evolves across updated decision steps as the task progresses. The manifestation of a historical step in future decision contexts is not fixed. Raw messages may become a compressed abstraction or even be dropped as irrelevant in the next decision step. Conversely, information that has been compressed into an abstraction or dropped may be reinstated as raw in the next step.

Furthermore, viewing from bottom to top along the vertical axis reveals, within a single decision step, the elastic type composition of information ranging from the earliest steps to the most recent steps. Evidently, the agent neither demands that all steps be raw messages nor that all steps be compressed abstractions; in fact, some historical information may be entirely ignored by the current decision step. Crucially, this pattern of multiple elastic types is not fixed but evolves over time. Additionally, we observe that the proportion of non‑drop information also fluctuates across different decision steps, which further corroborates the slowly fluctuating growth trend of the per‑step input token length of ACE reported in Sec.~\ref{sec:ctx_len}. 

These findings underscore the necessity of adaptively organizing the context in accordance with the task state as the agent proceeds. By contrast, existing approaches rigidly employ truncation or summarization solely for reducing context length, while neglecting the organization of effective context.

\subsection{Ablation Study on the Composition of Elastic Types}
\label{sec:elas_type}
To further analyze the impact of different elastic type compositions on ACE’s performance, we compared three additional combinations, including $\{\text{Raw},\,\text{Abs}\}$, $\{\text{Raw},\,\text{Drop}\}$, and $\{\text{Abs},\,\text{Drop}\}$, under the setting where ACE is adapted to the ReAct framework. The comparison results on GAIA are presented in Table~\ref{tab:elas_type}.

It is apparent that any of the compared combinations leads to a performance degradation of ACE. Among them, although $\{\text{Raw},\,\text{Abs}\}$ minimizes information loss, it inevitably introduces task‑irrelevant redundancy that hampers effective reasoning and decision making, thus leading to performance drops. As for $\{\text{Raw},\,\text{Drop}\}$, it is similar to truncation in that each historical step can only be either raw messages or dropped. Nevertheless, ACE with $\{\text{Raw},\,\text{Drop}\}$ still possesses a flexibility edge over truncation, yielding a modest overall advantage. Specifically, when GPT‑4.1 is used as the main LLM, the average performance is 37.6, compared with 37.0 for truncation in Table~\ref{tab:bsl_comp}; and when Gemini-3.1-flash-lite-preview is used as the main LLM, it achieves 44.8, compared with 43.0 for truncation in Table~\ref{tab:bsl_comp}.

In addition, the $\{\text{Abs},\,\text{Drop}\}$ setting is associated with the greatest degree of information loss. With this configuration, the agent is perpetually unable to extract adequate detail from raw messages, and is forced to perform reasoning and decision making based solely on abstracted information. Consequently, its final judgments are severely affected, leading to the lowest average performance among all compared settings.
\begin{wraptable}{r}{0.48\textwidth}
\centering
\vspace{-5pt}
\caption{Ablation study on the elastic type compositions under the setting of adapting ACE to ReAct.}
\label{tab:elas_type}
\begin{tabular}{p{2.8cm}<{\centering} | p{0.7cm}<{\centering} p{0.7cm}<{\centering} p{0.7cm}<{\centering} p{0.7cm}<{\centering}}
\toprule
\multirow{2}{*}{\textbf{Benchmark}} & \multicolumn{4}{c}{\textbf{GAIA}}  \\
& Text & MM & File & All \\
\midrule
\multicolumn{5}{c}{\textit{Main LLM: GPT-4.1}} \\
\midrule
$\{\text{Raw},\,\text{Abs}\}$  & 43.7 & 29.2 & 23.7 & 37.0   \\
$\{\text{Raw},\,\text{Drop}\}$ & 47.6 & 20.8 & 21.1 & 37.6  \\
$\{\text{Abs},\,\text{Drop}\}$ & 38.8 & 12.5 & 5.3  & 27.3  \\
\textbf{$\{\text{Raw},\,\text{Abs},\,\text{Drop}\}$} & \textbf{51.5} & \textbf{33.3} & \textbf{23.7} & \textbf{42.4}  \\
\midrule
\multicolumn{5}{c}{\textit{Main LLM: Gemini-3.1-flash-lite-preview}} \\
\midrule
$\{\text{Raw},\,\text{Abs}\}$  & 58.3 & 37.5 & 23.7 & 47.3 \\
$\{\text{Raw},\,\text{Drop}\}$ & 53.4 & 41.7 & 23.7 & 44.8  \\
$\{\text{Abs},\,\text{Drop}\}$ & 47.6 & 33.3 & 13.2 & 37.6  \\
\textbf{$\{\text{Raw},\,\text{Abs},\,\text{Drop}\}$} & \textbf{62.1} & \textbf{50.0} & \textbf{28.9} & \textbf{52.7} \\
\bottomrule
\end{tabular}
\caption{Ablation study on the elastic type of last step under the ReAct-based ACE setup.}
\label{tab:last_step_abla}
\begin{tabular}{p{2.8cm}<{\centering} | p{0.7cm}<{\centering} p{0.7cm}<{\centering} p{0.7cm}<{\centering} p{0.7cm}<{\centering}}
\toprule
\multirow{2}{*}{\textbf{Benchmark}} & \multicolumn{4}{c}{\textbf{GAIA}}  \\
& Text & MM & File & All \\
\midrule
\multicolumn{5}{c}{\textit{Main LLM: GPT-4.1}} \\
\midrule
Adaptive & 49.5 & \textbf{37.5} & 21.1 & 41.2  \\
\textbf{Kept Raw} & \textbf{51.5} & 33.3 & \textbf{23.7} & \textbf{42.4}  \\
\midrule
\multicolumn{5}{c}{\textit{Main LLM: Gemini-3.1-flash-lite-preview}} \\
\midrule
Adaptive & 57.3 & 50.0 & 26.3 & 49.1 \\
\textbf{Kept Raw}         & \textbf{62.1} & \textbf{50.0} & \textbf{28.9} & \textbf{52.7} \\
\bottomrule
\end{tabular}
\vspace{-40pt}
\end{wraptable}

It is imperative for agent context management to adaptively orchestrate historical step information across the three distinct elastic types in accordance with task progression. The omission of any single type inevitably incurs either accumulation of redundant information or excessive loss of detail, leading to performance deterioration. These observations further corroborate the validity of the proposed ACE framework.

\subsection{Ablation Study on the Elastic Type of Last Step}
\label{sec:last_step}
Another important design feature in ACE is that we always retain the information of the most recent step as raw messages when elastically orchestrating historical steps to build the context for each decision step. This ensures that the latest and most complete task state is available. To assess the necessity of this design, we conduct a comparison on GAIA under the ReAct‑based ACE setup, evaluating two settings: one with adaptive orchestration of the last step, and the other with it  kept as raw. Table~\ref{tab:last_step_abla} summarizes the comparison results.

From the experimental results, we observe that preserving the information of the last step in its raw message form throughout the context leads to robust performance improvements. Conversely, allowing adaptive adjustments to the last step carries the risk of misrepresenting the latest task state, compelling the agent to reason from outdated states and thus adversely affecting task progress. Thus, the effectiveness of the design that retains the last step as raw is validated.

\section{Conclusions}
We propose the ACE, a training‑free and plug‑and‑play context elasticizer that addresses the inflexibility and irreversibility of existing agent context management methods. By decoupling lossless message storage from per‑step elastic orchestration, ACE allows historical information to be adaptively presented as raw, abstraction, or dropped according to the evolving task state, while preserving the ability to recover previously discarded or compressed content when needed. We integrate ACE into four diverse agent frameworks without modifying their native architectures, and demonstrate consistent performance improvements across GAIA, HLE, WebShop, and other benchmarks. 
\bibliographystyle{unsrtnat}
\bibliography{ref}

@inproceedings{yao2023reactsynergizingreasoningacting,
  title={ReAct: Synergizing Reasoning and Acting in Language Models},
  author={Yao, Shunyu and Zhao, Jeffrey and Yu, Dian and Du, Nan and Shafran, Izhak and Narasimhan, Karthik and Cao, Yuan},
  booktitle={International Conference on Learning Representations (ICLR)},
  year={2023}
}

@article{li2025webthinkerempoweringlargereasoning,
  title={Webthinker: Empowering large reasoning models with deep research capability},
  author={Li, Xiaoxi and Jin, Jiajie and Dong, Guanting and Qian, Hongjin and Wu, Yongkang and Wen, Ji-Rong and Zhu, Yutao and Dou, Zhicheng},
  journal={Advances in Neural Information Processing Systems},
  volume={38},
  pages={120091--120131},
  year={2026}
}

@inproceedings{li2026deepagentgeneralreasoningagent,
  title={Deepagent: A general reasoning agent with scalable toolsets},
  author={Li, Xiaoxi and Jiao, Wenxiang and Jin, Jiarui and Dong, Guanting and Jin, Jiajie and Wang, Yinuo and Wang, Hao and Zhu, Yutao and Wen, Ji-Rong and Lu, Yuan and others},
  booktitle={Proceedings of the ACM Web Conference 2026},
  pages={2219--2230},
  year={2026}
}

@article{su2026miroflowhighperformancerobustopensource,
  title={Miroflow: Towards high-performance and robust open-source agent framework for general deep research tasks},
  author={Su, Shiqian and Xing, Sen and Dong, Xuan and Zhong, Muyan and Wang, Bin and Zhu, Xizhou and Chen, Yuntao and Wang, Wenhai and Deng, Yue and Zhu, Pengxiang and others},
  journal={arXiv preprint arXiv:2602.22808},
  year={2026}
}

@article{kang2025aconoptimizingcontextcompression,
  title={Acon: Optimizing context compression for long-horizon llm agents},
  author={Kang, Minki and Chen, Wei-Ning and Han, Dongge and Inan, Huseyin A and Wutschitz, Lukas and Chen, Yanzhi and Sim, Robert and Rajmohan, Saravan},
  journal={arXiv preprint arXiv:2510.00615},
  year={2025}
}

@article{wu2026resumunlockinglonghorizonsearch,
  title={Resum: Unlocking long-horizon search intelligence via context summarization},
  author={Wu, Xixi and Li, Kuan and Zhao, Yida and Zhang, Liwen and Ou, Litu and Yin, Huifeng and Zhang, Zhongwang and Yu, Xinmiao and Zhang, Dingchu and Jiang, Yong and others},
  journal={arXiv preprint arXiv:2509.13313},
  year={2025}
}

@article{ye2025agentfoldlonghorizonwebagents,
  title={AgentFold: Long-Horizon Web Agents with Proactive Context Management},
  author={Ye, Rui and Zhang, Zhongwang and Li, Kuan and Yin, Huifeng and Tao, Zhengwei and Zhao, Yida and Su, Liangcai and Zhang, Liwen and Qiao, Zile and Wang, Xinyu and others},
  journal={arXiv preprint arXiv:2510.24699},
  year={2025}
}

@article{zhang2026memoryactionautonomouscontext,
  title={Memory as action: Autonomous context curation for long-horizon agentic tasks},
  author={Zhang, Yuxiang and Shu, Jiangming and Ma, Ye and Lin, Xueyuan and Wu, Shangxi and Sang, Jitao},
  journal={arXiv preprint arXiv:2510.12635},
  year={2025}
}

@inproceedings{zhou2025mem1learningsynergizememory,
  title={MEM1: Learning to Synergize Memory and Reasoning for Efficient Long-Horizon Agents},
  author={Zhou, Zijian and Qu, Ao and Wu, Zhaoxuan and Kim, Sunghwan and Prakash, Alok and Rus, Daniela and Zhao, Jinhua and Low, Bryan Kian Hsiang and Liang, Paul Pu},
  booktitle={First Workshop on Multi-Turn Interactions in Large Language Models}
}

@article{yu2025memagentreshapinglongcontextllm,
  title={Memagent: Reshaping long-context llm with multi-conv rl-based memory agent},
  author={Yu, Hongli and Chen, Tinghong and Feng, Jiangtao and Chen, Jiangjie and Dai, Weinan and Yu, Qiying and Zhang, Ya-Qin and Ma, Wei-Ying and Liu, Jingjing and Wang, Mingxuan and others},
  journal={arXiv preprint arXiv:2507.02259},
  year={2025}
}

@article{chen2026iterresearchrethinkinglonghorizonagents,
  title={IterResearch: Rethinking Long-Horizon Agents with Interaction Scaling},
  author={Chen, Guoxin and Qiao, Zile and Chen, Xuanzhong and Yu, Donglei and Xu, Haotian and Zhao, Wayne Xin and Song, Ruihua and Yin, Wenbiao and Yin, Huifeng and Zhang, Liwen and others},
  journal={arXiv preprint arXiv:2511.07327},
  year={2025}
}

@article{beltagy2020longformerlongdocumenttransformer,
  title={Longformer: The long-document transformer},
  author={Beltagy, Iz and Peters, Matthew E and Cohan, Arman},
  journal={arXiv preprint arXiv:2004.05150},
  year={2020}
}

@article{zaheer2021bigbirdtransformerslonger,
  title={Big bird: Transformers for longer sequences},
  author={Zaheer, Manzil and Guruganesh, Guru and Dubey, Kumar Avinava and Ainslie, Joshua and Alberti, Chris and Ontanon, Santiago and Pham, Philip and Ravula, Anirudh and Wang, Qifan and Yang, Li and others},
  journal={Advances in neural information processing systems},
  volume={33},
  pages={17283--17297},
  year={2020}
}

@article{chen2023extendingcontextwindowlarge,
  title={Extending context window of large language models via positional interpolation},
  author={Chen, Shouyuan and Wong, Sherman and Chen, Liangjian and Tian, Yuandong},
  journal={arXiv preprint arXiv:2306.15595},
  year={2023}
}

@article{wang2026autoagent,
  title={Autoagent: Evolving cognition and elastic memory orchestration for adaptive agents},
  author={Wang, Xiaoxing and Liao, Ning and Wei, Shikun and Tang, Chen and Xiong, Feiyu},
  journal={arXiv preprint arXiv:2603.09716},
  year={2026}
}

@inproceedings{peng2026yarnefficientcontextwindow,
  title={Yarn: Efficient context window extension of large language models},
  author={Peng, Bowen and Quesnelle, Jeffrey and Fan, Honglu and Shippole, Enrico},
  booktitle={International Conference on Learning Representations},
  volume={2024},
  pages={31932--31951},
  year={2024}
}

@article{laban2025llmslostmultiturnconversation,
  title={Llms get lost in multi-turn conversation},
  author={Laban, Philippe and Hayashi, Hiroaki and Zhou, Yingbo and Neville, Jennifer},
  journal={arXiv preprint arXiv:2505.06120},
  year={2025}
}

@article{liu2023lostmiddlelanguagemodels,
  title={Lost in the middle: How language models use long contexts},
  author={Liu, Nelson F and Lin, Kevin and Hewitt, John and Paranjape, Ashwin and Bevilacqua, Michele and Petroni, Fabio and Liang, Percy},
  journal={Transactions of the association for computational linguistics},
  volume={12},
  pages={157--173},
  year={2024}
}

@inproceedings{yuan2025nativesparseattentionhardwarealigned,
  title={Native sparse attention: Hardware-aligned and natively trainable sparse attention},
  author={Yuan, Jingyang and Gao, Huazuo and Dai, Damai and Luo, Junyu and Zhao, Liang and Zhang, Zhengyan and Xie, Zhenda and Wei, Yuxing and Wang, Lean and Xiao, Zhiping and others},
  booktitle={Proceedings of the 63rd Annual Meeting of the Association for Computational Linguistics (Volume 1: Long Papers)},
  pages={23078--23097},
  year={2025}
}

@article{deepseekai2025deepseekv32pushingfrontieropen,
  title={Deepseek-v3. 2: Pushing the frontier of open large language models},
  author={Liu, Aixin and Mei, Aoxue and Lin, Bangcai and Xue, Bing and Wang, Bingxuan and Xu, Bingzheng and Wu, Bochao and Zhang, Bowei and Lin, Chaofan and Dong, Chen and others},
  journal={arXiv preprint arXiv:2512.02556},
  year={2025}
}

@inproceedings{bai2024longbenchbilingualmultitaskbenchmark,
  title={Longbench: A bilingual, multitask benchmark for long context understanding},
  author={Bai, Yushi and Lv, Xin and Zhang, Jiajie and Lyu, Hongchang and Tang, Jiankai and Huang, Zhidian and Du, Zhengxiao and Liu, Xiao and Zeng, Aohan and Hou, Lei and others},
  booktitle={Proceedings of the 62nd annual meeting of the association for computational linguistics (volume 1: Long papers)},
  pages={3119--3137},
  year={2024}
}

@article{zeng2026glm,
  title={Glm-5: from vibe coding to agentic engineering},
  author={Zeng, Aohan and Lv, Xin and Hou, Zhenyu and Du, Zhengxiao and Zheng, Qinkai and Chen, Bin and Yin, Da and Ge, Chendi and Huang, Chenghua and Xie, Chengxing and others},
  journal={arXiv preprint arXiv:2602.15763},
  year={2026}
}

@article{team2025kimi,
  title={Kimi k2: Open agentic intelligence},
  author={Team, Kimi and Bai, Yifan and Bao, Yiping and Charles, Y and Chen, Cheng and Chen, Guanduo and Chen, Haiting and Chen, Huarong and Chen, Jiahao and Chen, Ningxin and others},
  journal={arXiv preprint arXiv:2507.20534},
  year={2025}
}

@article{chen2026minimax,
  title={The minimax-m2 series: Mini activations unleashing max real-world intelligence},
  author={Chen, Aili and Li, Aonian and Zhou, Baichuan and Gong, Bangwei and Jiang, Binyang and Dan, Boji and Yu, Changqing and Wang, Chao and Ma, Cheng and Zhong, Cheng and others},
  journal={arXiv preprint arXiv:2605.26494},
  year={2026}
}

@article{team2025longcat,
  title={Longcat-flash technical report},
  author={Team, Meituan LongCat and Li, Bei and Lei, Bingye and Wang, Bo and Rong, Bolin and Wang, Chao and Zhang, Chao and Gao, Chen and Zhang, Chen and Sun, Cheng and others},
  journal={arXiv preprint arXiv:2509.01322},
  year={2025}
}

@article{Wang_2024,
  title={A survey on large language model based autonomous agents},
  author={Wang, Lei and Ma, Chen and Feng, Xueyang and Zhang, Zeyu and Yang, Hao and Zhang, Jingsen and Chen, Zhiyuan and Tang, Jiakai and Chen, Xu and Lin, Yankai and others},
  journal={Frontiers of Computer Science},
  volume={18},
  number={6},
  pages={186345},
  year={2024},
  publisher={Springer}
}

@article{shinn2023reflexionlanguageagentsverbal,
  title={Reflexion: Language agents with verbal reinforcement learning},
  author={Shinn, Noah and Cassano, Federico and Gopinath, Ashwin and Narasimhan, Karthik and Yao, Shunyu},
  journal={Advances in neural information processing systems},
  volume={36},
  pages={8634--8652},
  year={2023}
}

@article{schick2023toolformerlanguagemodelsteach,
  title={Toolformer: Language models can teach themselves to use tools},
  author={Schick, Timo and Dwivedi-Yu, Jane and Dess{\`\i}, Roberto and Raileanu, Roberta and Lomeli, Maria and Hambro, Eric and Zettlemoyer, Luke and Cancedda, Nicola and Scialom, Thomas},
  journal={Advances in neural information processing systems},
  volume={36},
  pages={68539--68551},
  year={2023}
}

@article{nakano2022webgptbrowserassistedquestionansweringhuman,
  title={Webgpt: Browser-assisted question-answering with human feedback},
  author={Nakano, Reiichiro and Hilton, Jacob and Balaji, Suchir and Wu, Jeff and Ouyang, Long and Kim, Christina and Hesse, Christopher and Jain, Shantanu and Kosaraju, Vineet and Saunders, William and others},
  journal={arXiv preprint arXiv:2112.09332},
  year={2021}
}

@inproceedings{jin2025searchr1trainingllmsreason,
  title={Search-R1: Training LLMs to Reason and Leverage Search Engines with Reinforcement Learning},
  author={Jin, Bowen and Zeng, Hansi and Yue, Zhenrui and Yoon, Jinsung and Arik, Sercan O and Wang, Dong and Zamani, Hamed and Han, Jiawei},
  booktitle={Second Conference on Language Modeling}
}

@article{li2025websailornavigatingsuperhumanreasoning,
  title={Websailor: Navigating super-human reasoning for web agent},
  author={Li, Kuan and Zhang, Zhongwang and Yin, Huifeng and Zhang, Liwen and Ou, Litu and Wu, Jialong and Yin, Wenbiao and Li, Baixuan and Tao, Zhengwei and Wang, Xinyu and others},
  journal={arXiv preprint arXiv:2507.02592},
  year={2025}
}

@article{zhang2024surveymemorymechanismlarge,
  title={A survey on the memory mechanism of large language model-based agents},
  author={Zhang, Zeyu and Dai, Quanyu and Bo, Xiaohe and Ma, Chen and Li, Rui and Chen, Xu and Zhu, Jieming and Dong, Zhenhua and Wen, Ji-Rong},
  journal={ACM Transactions on Information Systems},
  volume={43},
  number={6},
  pages={1--47},
  year={2025},
  publisher={ACM New York, NY}
}

@article{hu2026memoryageaiagents,
  title={Memory in the age of ai agents},
  author={Hu, Yuyang and Liu, Shichun and Yue, Yanwei and Zhang, Guibin and Liu, Boyang and Zhu, Fangyi and Lin, Jiahang and Guo, Honglin and Dou, Shihan and Xi, Zhiheng and others},
  journal={arXiv preprint arXiv:2512.13564},
  year={2025}
}

@article{sun2025scalinglonghorizonllmagent,
  title={Scaling long-horizon llm agent via context-folding},
  author={Sun, Weiwei and Lu, Miao and Ling, Zhan and Liu, Kang and Yao, Xuesong and Yang, Yiming and Chen, Jiecao},
  journal={arXiv preprint arXiv:2510.11967},
  year={2025}
}

@article{qian2026memobrainexecutivememoryagentic,
  title={MemoBrain: Executive Memory as an Agentic Brain for Reasoning},
  author={Qian, Hongjin and Cao, Zhao and Liu, Zheng},
  journal={arXiv preprint arXiv:2601.08079},
  year={2026}
}

@article{du2026memoryautonomousllmagentsmechanisms,
  title={Memory for autonomous llm agents: Mechanisms, evaluation, and emerging frontiers},
  author={Du, Pengfei},
  journal={arXiv preprint arXiv:2603.07670},
  year={2026}
}

@article{liang2026genericagenttokenefficientselfevolvingllm,
  title={GenericAgent: A Token-Efficient Self-Evolving LLM Agent via Contextual Information Density Maximization (V1. 0)},
  author={Liang, Jiaqing and Han, Jinyi and Li, Weijia and Wang, Xinyi and Zhang, Zhoujia and Jiang, Zishang and Liao, Ying and Li, Tingyun and Huang, Ying and Shen, Hao and others},
  journal={arXiv preprint arXiv:2604.17091},
  year={2026}
}

@article{miromindteam2026mirothinker17h1heavyduty,
  title={Mirothinker-1.7 \& h1: Towards heavy-duty research agents via verification},
  author={Team, MiroMind and Bai, S and Bing, L and Lei, L and Li, R and Li, X and Lin, X and Min, E and Su, L and Wang, B and others},
  journal={arXiv preprint arXiv:2603.15726},
  year={2026}
}

@article{yao2026arcactivereflectiondrivencontext,
  title={ARC: Active and Reflection-driven Context Management for Long-Horizon Information Seeking Agents},
  author={Yao, Yilun and Huang, Shan and Dai, Elsie and Tan, Zhewen and Duan, Zhenyu and Jia, Shousheng and Jiang, Yanbing and Yang, Tong},
  journal={arXiv preprint arXiv:2601.12030},
  year={2026}
}

@inproceedings{mialon2023gaiabenchmarkgeneralai,
  title={Gaia: a benchmark for general ai assistants},
  author={Mialon, Gr{\'e}goire and Fourrier, Cl{\'e}mentine and Wolf, Thomas and LeCun, Yann and Scialom, Thomas},
  booktitle={International Conference on Learning Representations},
  volume={2024},
  pages={9025--9049},
  year={2024}
}

@article{phan2025humanity,
  title={Humanity's last exam},
  author={Phan, Long and Gatti, Alice and Han, Ziwen and Li, Nathaniel and Hu, Josephina and Zhang, Hugh and Zhang, Chen Bo Calvin and Shaaban, Mohamed and Ling, John and Shi, Sean and others},
  journal={arXiv preprint arXiv:2501.14249},
  year={2025}
}

@article{yao2023webshopscalablerealworldweb,
  title={Webshop: Towards scalable real-world web interaction with grounded language agents},
  author={Yao, Shunyu and Chen, Howard and Yang, John and Narasimhan, Karthik},
  journal={Advances in Neural Information Processing Systems},
  volume={35},
  pages={20744--20757},
  year={2022}
}

@inproceedings{wu2025webwalkerbenchmarkingllmsweb,
  title={Webwalker: Benchmarking llms in web traversal},
  author={Wu, Jialong and Yin, Wenbiao and Jiang, Yong and Wang, Zhenglin and Xi, Zekun and Fang, Runnan and Zhang, Linhai and He, Yulan and Zhou, Deyu and Xie, Pengjun and others},
  booktitle={Proceedings of the 63rd Annual Meeting of the Association for Computational Linguistics (Volume 1: Long Papers)},
  pages={10290--10305},
  year={2025}
}

@article{chen2025xbenchtrackingagentsproductivity,
  title={xbench: Tracking agents productivity scaling with profession-aligned real-world evaluations},
  author={Chen, Kaiyuan and Ren, Yixin and Liu, Yang and Hu, Xiaobo and Tian, Haotong and Xie, Tianbao and Liu, Fangfu and Zhang, Haoye and Liu, Hongzhang and Gong, Yuan and others},
  journal={arXiv preprint arXiv:2506.13651},
  year={2025}
}

@article{zhou2025browsecompzhbenchmarkingwebbrowsing,
  title={Browsecomp-zh: Benchmarking web browsing ability of large language models in chinese},
  author={Zhou, Peilin and Leon, Bruce and Ying, Xiang and Zhang, Can and Shao, Yifan and Ye, Qichen and Chong, Dading and Jin, Zhiling and Xie, Chenxuan and Cao, Meng and others},
  journal={arXiv preprint arXiv:2504.19314},
  year={2025}
}



\end{document}